\documentclass[10pt,twocolumn,letterpaper]{article}

\usepackage[pagenumbers]{cvpr} 
\usepackage{marvosym}
\usepackage{times}
\usepackage{epsfig}
\usepackage{graphicx}
\usepackage{amsmath}
\usepackage{amssymb}
\usepackage{array}
\usepackage{makecell}
\usepackage{multirow}
\usepackage{threeparttable}
\usepackage{colortbl}
\definecolor{shadow}{cmyk}{.08,0,0,0}

\usepackage[pagebackref=true,breaklinks=true,letterpaper=true,colorlinks,bookmarks=false]{hyperref}

\def\cvprPaperID{6384} 
\def\confName{CVPR}
\def\confYear{2022}

\begin{document}

\title{Occlusion-Aware Cost Constructor for Light Field Depth Estimation}

\author{Yingqian Wang, Longguang Wang, Zhengyu Liang, Jungang Yang, Wei An, Yulan Guo\\
National University of Defense Technology\\
\tt\small \url{https://github.com/YingqianWang/OACC-Net}}

\maketitle

\begin{abstract}
Matching cost construction is a key step in light field (LF) depth estimation, but was rarely studied in the deep learning era. Recent deep learning-based LF depth estimation methods construct matching cost by sequentially shifting each sub-aperture image (SAI) with a series of predefined offsets, which is complex and time-consuming. In this paper, we propose a simple and fast cost constructor to construct matching cost for LF depth estimation. Our cost constructor is composed by a series of convolutions with specifically designed dilation rates. By applying our cost constructor to SAI arrays, pixels under predefined disparities can be integrated and matching cost can be constructed without using any shifting operation. More importantly, the proposed cost constructor is occlusion-aware and can handle occlusions by dynamically modulating pixels from different views. Based on the proposed cost constructor, we develop a deep network for LF depth estimation. Our network ranks first on the commonly used 4D LF benchmark in terms of the mean square error (MSE), and achieves a faster running time than other state-of-the-art methods.

\end{abstract}

\section{Introduction}
Light field (LF) cameras can encode 3D scenes into 4D LF images. By using the abundant spatial and angular information in the LF images, the scene depth can be obtained by performing LF depth estimation. As a fundamental task in LF image processing, depth estimation benefits many subsequent applications such as refocusing \cite{wang2018selective}, view synthesis \cite{FS-GAF,guo2021learning}, 3D reconstruction \cite{kim2013scene}, and virtual reality \cite{yu2017light}.

With the advances of deep neural networks, many deep learning-based methods \cite{heber2016convolutional,heber2017neural,EPINET,EPI-Shift,ORM,LFAttNet,AttMLFNet,FastLFnet} have been proposed and boosted the performance of LF depth estimation. Recent deep learning-based methods achieve LF depth estimation in a four-step pipeline including feature extraction, cost construction, cost aggregation, and depth regression. To achieve higher accuracy, these methods designed different modules for feature extraction \cite{LFAttNet} and cost aggregation \cite{AttMLFNet,FastLFnet}. However, as a key step in LF depth estimation, matching cost construction was rarely studied.

\begin{figure}[t]
\centering
\includegraphics[width=8cm]{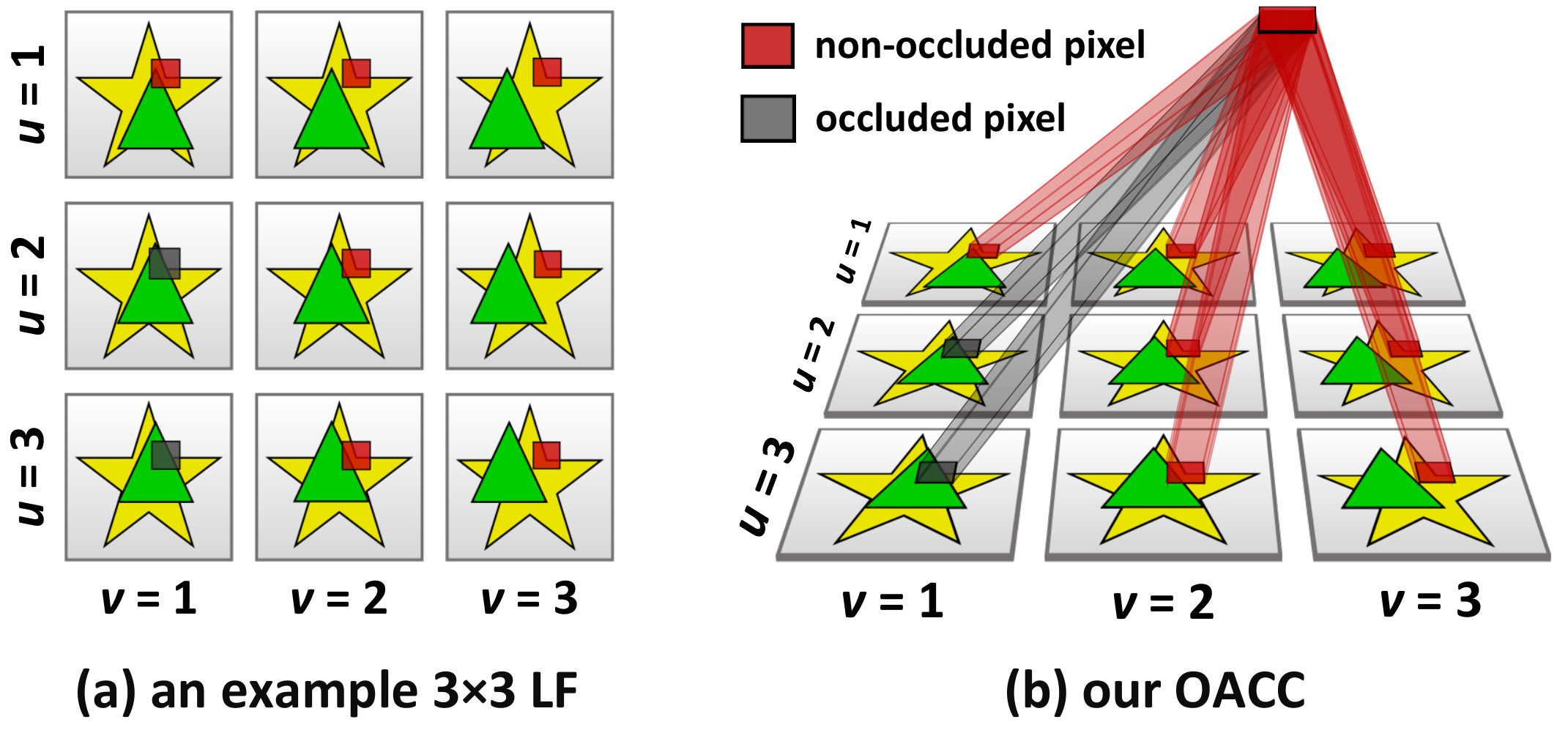}
\vspace{-0.1cm}
\caption{An illustration of the proposed occlusion-aware cost constructor (OACC). (a) A toy example of a 3$\times$3 LF, in which the yellow star is partially occluded by the green triangle. (b) Our OACC constructs matching cost via convolutions and can handle occlusions by assigning smaller weights to occluded pixels.} \label{fig:thumbnail}
\vspace{-0.2cm}
\end{figure}

To construct matching costs for LF depth estimation, existing methods \cite{LFAttNet,AttMLFNet} shift each sub-aperture image (SAI) with a series of predefined offsets, and then concatenate the shifted SAIs to form a cost volume. Although this \textit{shift-and-concat} scheme is easy to implement, the large number of shifting operation\footnote{For example, in \textit{LFAttNet} \cite{LFAttNet}, totally 80 views are shifted by 8 disparity levels, resulting in 640 sequential shifting operations.} reduces the efficiency of these methods. Moreover, during matching cost construction, pixels at different spatial locations are processed equally, which cannot handle the spatially-varying occlusions where some views are less informative and can even deteriorate the estimation results.

To handle the aforementioned challenges, in this paper, we propose an occlusion-aware cost constructor (OACC) for LF depth estimation. Our cost constructor is composed by a series of convolutions with specifically designed dilation rates. By applying our OACC to SAI arrays, pixels under predefined disparities can be integrated without performing shifting operation. More importantly, our OACC can handle occlusions by dynamically modulating pixels from different views, as shown in Fig.~\ref{fig:thumbnail}. Based on the proposed OACC, we develop a deep network for LF depth estimation. Our network achieves state-of-the-art depth estimation accuracy with a significant acceleration.

The contributions of this paper can be summarized as:
\begin{itemize}
    \item We propose a cost constructor to replace the \textit{shift-and-concat} approach for matching cost construction.
    \item We make our cost constructor to be occlusion-aware by modulating pixels from different views in a fine-grained manner.
    \item We develope an OACC-Net for LF depth estimation. Our method achieves top accuracy with significant acceleration as compared to other state-of-the-art methods on the 4D LF benchmark \cite{HCInew}.
\end{itemize}

\section{Related Works}
In this section, we review the major works in LF depth estimation. We classify the existing methods into traditional methods and deep learning-based methods.

\subsection{Traditional Methods}
Early works on LF depth estimation follow the traditional paradigm and use different approaches to measure the consistency among different views. Tao et al. \cite{tao2013depth} proposed to combine the correspondence cue and the defocus cue for LF depth estimation. Subsequently, Tao et al. \cite{tao2015depth} introduced a shading-based refinement approach to improve the depth estimation accuracy. Jeon et al. \cite{jeon2015accurate} proposed a phase-based multi-view stereo matching method and achieved depth estimation in the Fourier domain. Wang et al. \cite{LF-OCC} considered occlusions in LF depth estimation and proposed an occlusion-aware algorithm based on the partial angular consistency.
Williem et al. \cite{CAE} proposed angular entropy cost and adaptive defocus cost to handle the noise and occlusion issues for depth estimation. More recently, Han et al. \cite{OAVC} proposed an occlusion-aware vote cost to preserve edges in depth maps.

Since an epipolar plane image (EPI) contains patterns of oriented lines and the slope of these lines is related to the depth values, many methods achieve depth estimation by analyzing the slope of each line on EPIs. Wanner et al. \cite{wanner2013variational} proposed a structure tensor to estimate the slope of lines in horizontal and vertical EPIs, and refined the initial results by global optimization. Zhang et al. \cite{SPO} proposed a spinning parallelogram operator (SPO) to estimate the slopes for depth estimation. Sheng et al. \cite{SPO-MO} proposed to estimate slopes using multi-orientation EPIs and achieved improved results over SPO. Schilling et al. \cite{OBER} proposed an inline occlusion handling scheme operated on EPIs to achieve state-of-the-art depth estimation performance among traditional methods.

\subsection{Deep Learning-based Methods}
Recently, deep networks have been widely used for depth estimation and achieved significant performance gain over traditional methods. Heber et al. \cite{heber2016convolutional} proposed the first end-to-end network to learn the mapping between a 4D LF and its corresponding depths. Subsequently, Heber et al. \cite{heber2017neural} proposed a U-shaped network with 3D convolutions to extract geometric information from LFs for depth estimation. Shin et al. \cite{EPINET} proposed a multi-stream network and a series of data augmentation approaches for fast and accurate LF depth estimation. Tsai et al. \cite{LFAttNet} proposed an attention-based view selection network to adaptively incorporate all angular views for depth estimation. Peng et al. \cite{peng2018unsupervised} proposed an unsupervised LF depth estimation method that can be trained without using the ground-truth depth maps. Subsequently, Peng et al. \cite{peng2020zero} proposed a zero-shot learning-based method that can perform unsupervised depth estimation without using external datasets. More recently, Chen et al. \cite{AttMLFNet} proposed an attention-based multi-level fusion network to handle the occlusion problem for depth estimation. Huang et al. \cite{FastLFnet} proposed a multi-disparity-scale cost aggregation approach to achieve fast LF depth estimation.

Different from existing methods which focus on designing advanced modules for feature extraction \cite{LFAttNet} or cost aggregation \cite{AttMLFNet,FastLFnet}, in this paper, we study the cost construction stage and propose a simple and efficient module to achieve occlusion-aware cost construction.

\section{Method}
 In this section, we first describe the LF structure and analyze the influence of occlusions to angular consistency. Then, we introduce the proposed occlusion-aware cost constructor. Finally, we introduce our network for LF depth estimation.

\subsection{LF Structure and Occlusion Analysis} \label{sec:LFstructure}

\begin{figure}[t]
 \centering
 \includegraphics[width=8cm]{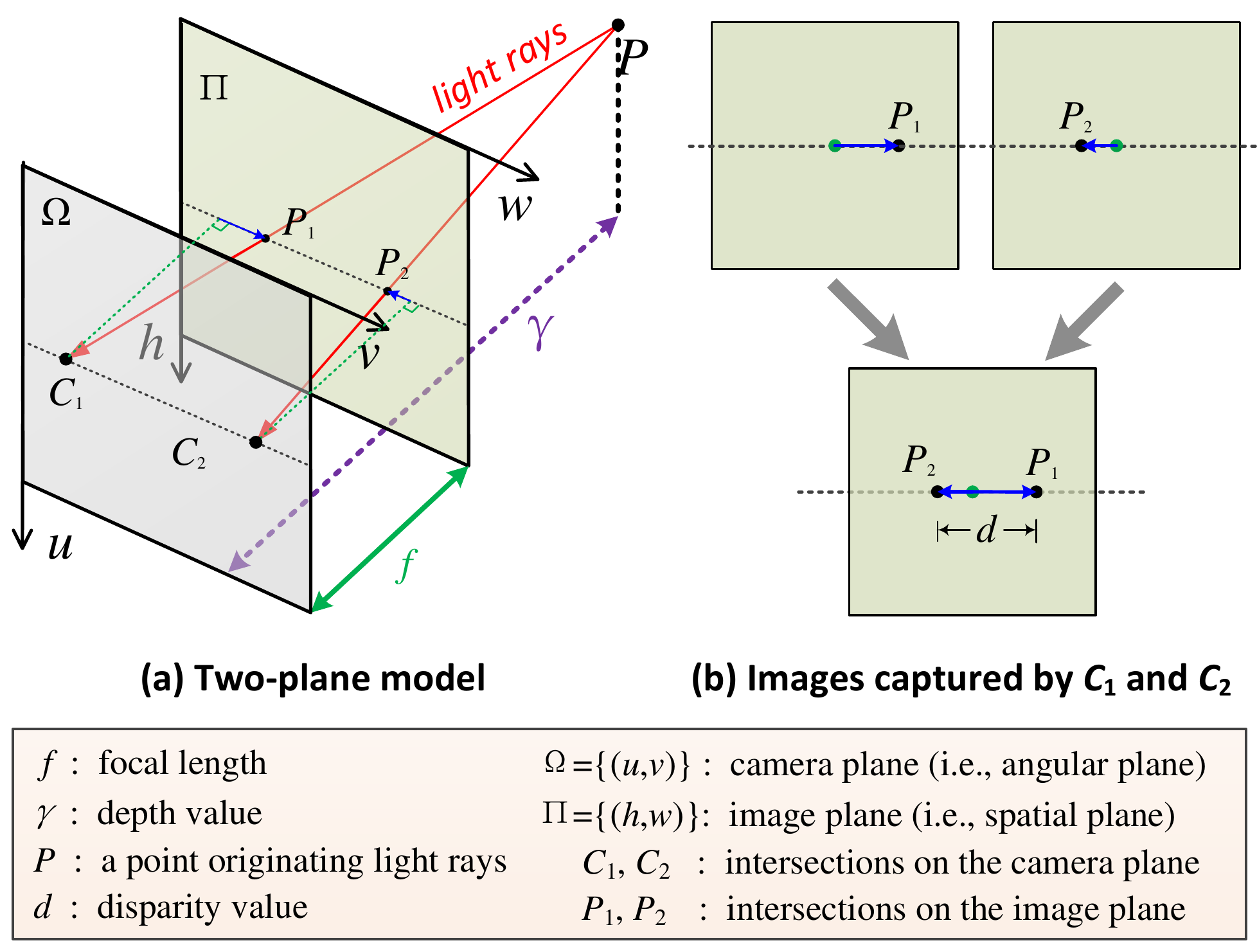}
 \vspace{-0.2cm}
 \caption{An illustration of the two-plane model.}  \label{fig:twoplane}
 \vspace{-0.2cm}
 \end{figure}

 We use the two-plane model \cite{levoy1996light} to parameterize LFs. As shown in Fig.~\ref{fig:twoplane}, a light ray originated from point $P$ can be uniquely determined by its intersections across the camera plane $\Omega=\{(u,v)\}$ and the image plane $\Pi=\{(h,w)\}$. Consequently, an LF can be formulated as a 4D tensor according to the mapping function $\mathcal{L}(u,v,h,w): \Omega \times \Pi \mapsto \mathbb{R}$. In this paper, we denote the 4D LF as $\mathcal{L}\in\mathbb{R^{\mathrm{\mathit{U}\times\mathit{V}\times\mathit{H}\times\mathit{W}}}}$, where $\mathit{U}$ and $\mathit{V}$ represent the angular dimensions, and $\mathit{H}$ and $\mathit{W}$ represent the spatial dimensions.

 Note that, a scene point (e.g., point $P$ in Fig.~\ref{fig:twoplane}) will be projected to different locations on the images captured by different cameras. As shown in Fig.~\ref{fig:twoplane}(b), there is a disparity (denoted as $d$) between projections $P_1$ and $P_2$, and the depth value of $P$ can be calculated according to $\gamma = fB/d$, where $B$ and $f$ represent the baseline length and the focal length of the LF camera. Consequently, depth estimation can be achieved by estimating the per-pixel disparities of the LF images. In this paper, we follow \cite{EPINET,LFAttNet,AttMLFNet,FastLFnet} to estimate the disparity map of the center view.

 Since the projections of a scene point on different views should have identical intensity under Lambertian and non-occlusion assumptions, depth estimation can be achieved by choosing the disparity candidate with highest angular consistency. To compare the angular consistency under different disparities, we construct angular patches according to
 \begin{equation}
 \mathcal{A}^{p}_{d}(u,v) = \mathcal{L}\left(u,v,h+(u_c-u)d, w+(v_c-v)d\right),
 \end{equation}
 where ${A}^{p}_{d}$ is the angular patch at pixel $p(h,w)$ with disparity $d$, and $u_c$ and $v_c$ represent the angular coordinates of the center view.

\begin{figure}[t]
\centering
\includegraphics[width=8.3cm]{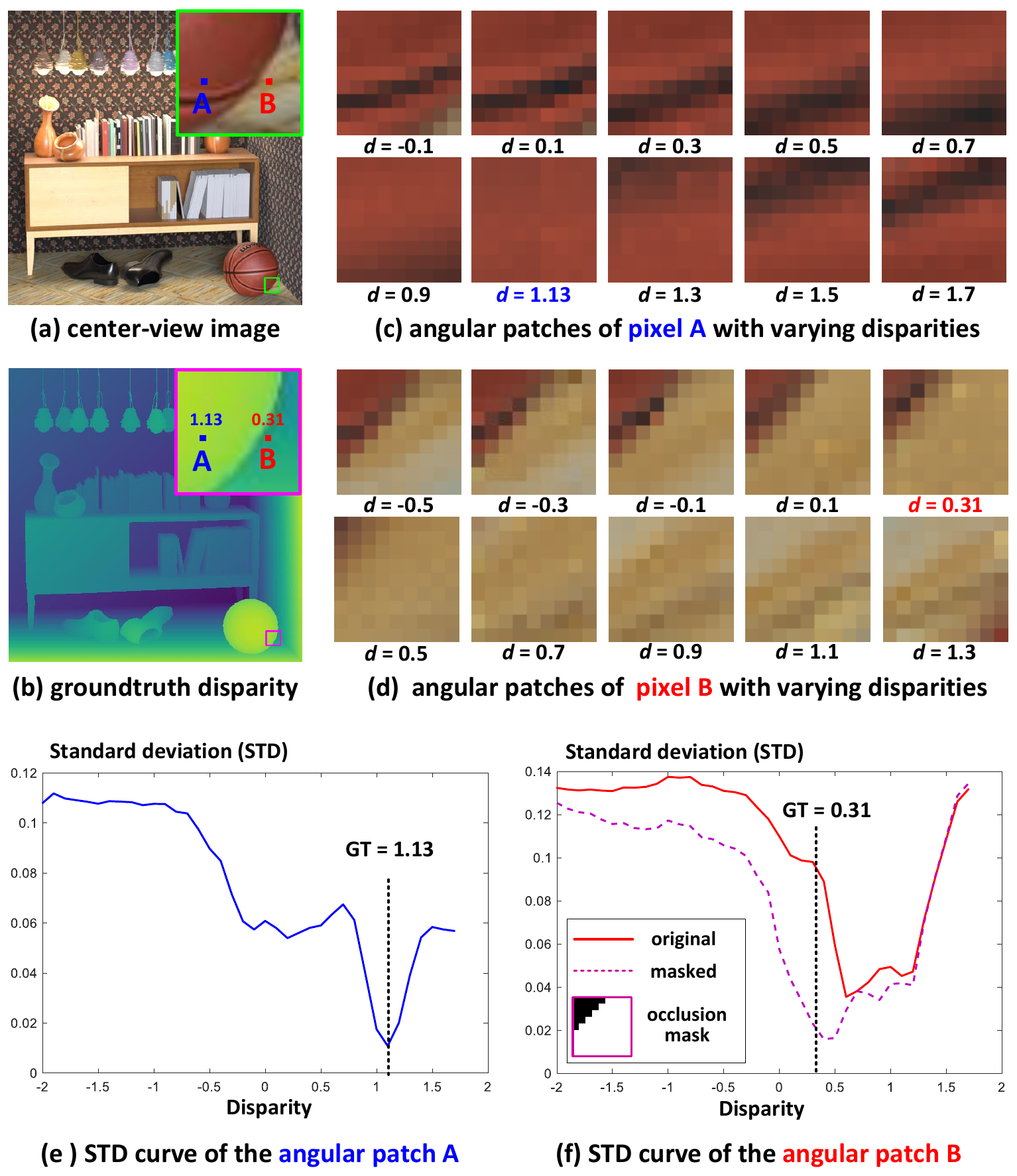}
\vspace{-0.5cm}
\caption{Comparison of the angular consistency in occluded and non-occluded regions. Standard deviation is used to evaluate the consistency of pixels in an angular patch. For the occluded region, an occlusion mask is needed to make the standard deviation reach its minimum at the correct disparity.} \label{fig:AngPatch}
\vspace{-0.2cm}
\end{figure}

 Here, we use an example for illustration. As shown in Fig. \ref{fig:AngPatch}(a), we select two pixels from scene \textit{sideboard} \cite{HCInew}, and construct angular patches under different disparities. The angular patches of pixel $\mathbf{A}$ are shown in Fig. \ref{fig:AngPatch}(c), from which we can see that the color of the pixels in the angular patch is more consistent near groundtruth disparity value (i.e., $d=1.13$). We further calculate the standard deviation of these angular patches to evaluate their consistency. As shown in Fig. \ref{fig:AngPatch}(e), the curve reaches its minimum at the groundtruth disparity. That is, the intensity of pixels in an angular patch is most consistent under correct disparities.

 However, this theory does not hold in occluded regions. As shown in Fig. \ref{fig:AngPatch}(d), we construct angular patches of pixel $\mathbf{B}$ under different disparities. It can be observed that pixels on the top-left corner of the angular patch have different color from other pixels even under the groundtruth disparity (i.e., $d=0.31$). That is because, the corresponding pixels in the top-left views are occluded by the foreground basketball. Consequently, occlusions can deteriorate the angular consistency under correct disparities, making the standard deviation curve (the red curve in Fig. \ref{fig:AngPatch}(f)) not reach its minimum at the groundtruth disparity.

 It is interesting that when we mask the top-left pixels in each angular patch of pixel $\mathbf{B}$ and calculate the standard deviation of the remaining pixels, the modified curve (the pink curve in Fig. \ref{fig:AngPatch}(f)) reaches its minimum near the groundtruth disparity. It demonstrates that the intensity of non-occluded pixels in an angular patch are most consistent at correct disparities. Motivated by this observation,  we design an occlusion-aware cost constructor to handle the occlusion issue for matching cost construction.

 \subsection{Occlusion-Aware Cost Constructor}
 Given a 5D LF feature $\mathcal{L}\in\mathbb{R}^{U \times V \times H \times W\times C}$ and a candidate disparity $d \in \{d_\textit{min},\cdots,d_\textit{max}\}$, existing methods construct matching costs using the \textit{shift-and-concat} approach. Specifically, they shift the feature of each view according to its angular coordinate $(u,v)$ and the given disparity $d$, then concatenate all the shifted features to generate the cost tensor. That is,
 \begin{equation}
 \mathcal{F}^{d}_{u,v}(h,w,:) = \mathcal{F}_{u,v}\left(h+(u_c\textit{--}u)d, w+(v_c\textit{--}v)d,:\right),
 \end{equation}
 \begin{equation}
 \mathcal{C}_{d} = \textit{Concat}\left(\mathcal{F}^{d}_{0,0}, \cdots, \mathcal{F}^{d}_{U,V}\right),
 \end{equation}
  where $\mathcal{F}^{d}_{u,v}\in \mathbb{R}^{H\times W\times C}$ denotes the shifted feature of view $(u,v)$ under disparity $d$, and $\mathcal{C}_{d}\in \mathbb{R}^{H\times W \times UVC}$ denotes the cost tensor under disparity $d$. Finally, the cost volume is generated by stacking all the cost tensors (i.e., $\mathcal{C}_{d_\textit{min}}, \cdots, \mathcal{C}_{d_\textit{max}}$) along the disparity dimension.

  Instead of using the \textit{shift-and-concat} approach, in this paper, we propose an occlusion-aware cost constructor for matching cost construction. The main ideas of our OACC are: 1) using convolutions to integrate pixels of each view under specific disparities; 2) modulating the input pixels to handle occlusions during cost construction.

 \begin{figure}
 \centering
 \includegraphics[width=8cm]{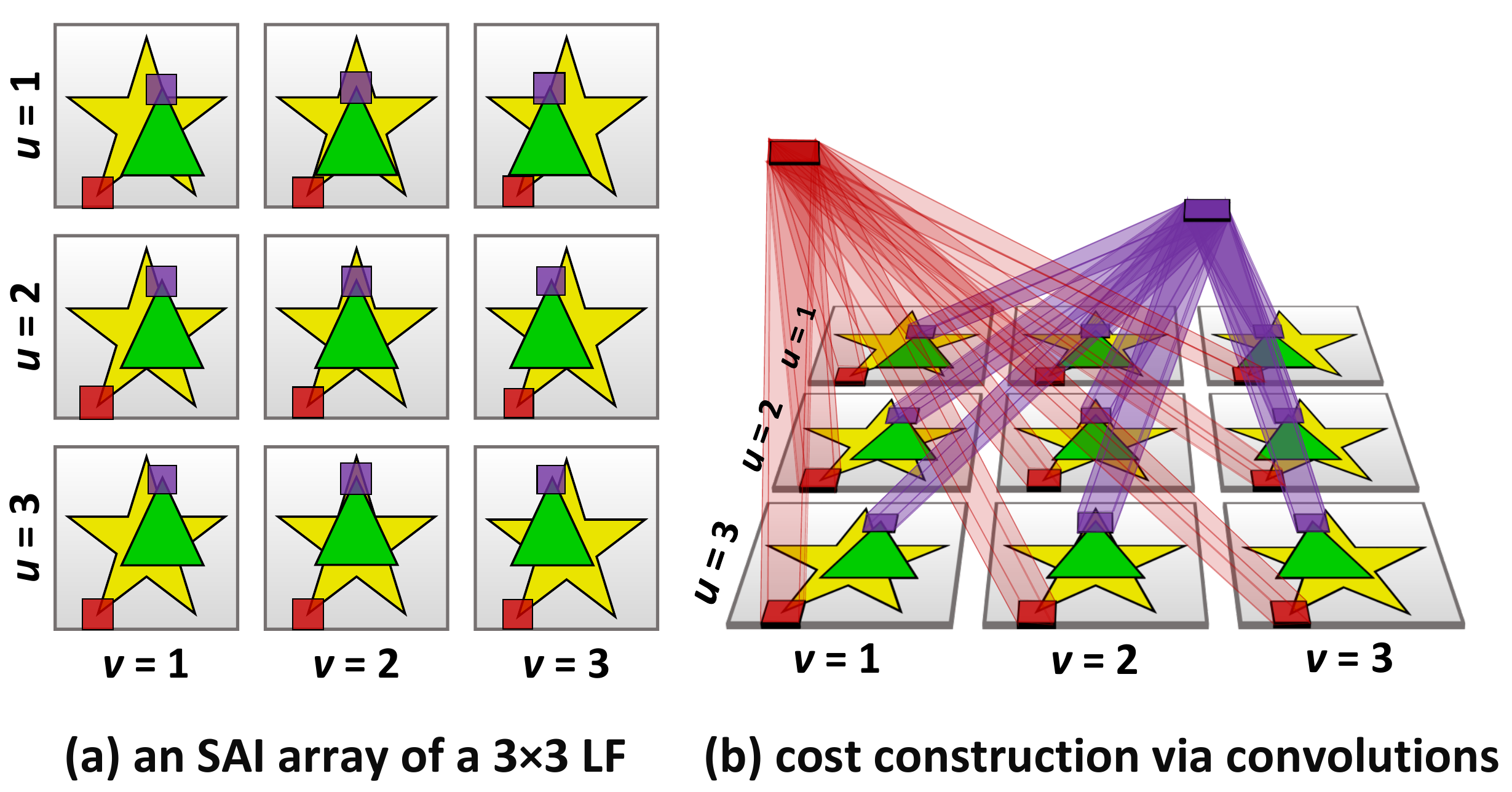}
 \vspace{-0.2cm}
 \caption{An illustration of our cost construction process.} \label{fig:DSAFE}
 \vspace{-0.3cm}
 \end{figure}

 \subsubsection{Cost Construction via Convolutions}\label{sec:CCviaConv}
 To construct matching cost, pixels in the angular patch under each candidate disparity should be integrated respectively. However, how to efficiently find the corresponding pixels to form an angular patch remains challenging. Since LF images have a regular spatial-angular structure \cite{LF-InterNet}, cost construction can be achieved by performing convolutions on the SAI array.

 Here, we use a toy example for illustration. As shown in Fig.~\ref{fig:DSAFE}(a), a 3$\times$3 LF is organized in an array of SAIs. In this scenario, the yellow star is in the background with zero disparity, and the green triangle is in the foreground and has a positive disparity\footnote{Under a positive disparity, object in the left$/$upper views locates at right$/$lower positions.}. The bottom-left vertex of the yellow star and the top vertex of the green triangle are marked by a red box and a purple box, respectively. It can be observed that both red and purple boxes are evenly distributed in a square region, which can be easily integrated via convolutions. Consequently, we design our cost constructor as a series of convolutions with a kernel size of $U$$\times$$V$ and different dilation rates to integrate angular patches under different disparities. The dilation rate is closely related to the preset disparity $d$ and can be calculated according to
 \begin{equation}\label{Eq:dilation}
 dila(d) = \left[H-d, W-d \right],
 \end{equation}
 where $H$ and $W$ denote the height and width of each SAI, respectively. From Eq.~\ref{Eq:dilation}, we can conclude that object with a larger disparity value (e.g., the purple box) has a smaller dilation rate, which is consistent with the toy example in Fig.~\ref{fig:DSAFE}. Using our proposed cost constructor, angular patches under different disparities can be integrated without performing any shifting operation, and  the matching cost can be efficiently constructed by convolving all the pixels in the angular patch, as shown  in Fig.~\ref{fig:DSAFE}(b).

  In our implementation, zero-padding is performed on each SAI to avoid aliasing among different views at boundaries. Moreover, the boundary of the resulting features are cropped to ensure the output features have a spatial resolution of $H$$\times$$W$. Details of padding and cropping strategies can be referred to the \href{https://yingqianwang.github.io/files/OACC-Net_supp.pdf}{supplemental material}.

 \subsubsection{Occlusion Handling via Pixel Modulation}
 As analyzed in Section \ref{sec:LFstructure}, occlusions can deteriorate the angular consistency and should be masked out during cost construction. Inspired by \textit{Deformable ConvNet V2} \cite{deformableV2}, we introduce a modulation mechanism to dynamically adjust the amplitude of pixels from different views for occlusion handling. Specifically, given an angular patch\footnote{Our modulated convolution is applied to the SAI arrays as in Fig.~\ref{fig:DSAFE}. For simplicity, we use the angular patch to denote the pixels from each view under a specific disparity, and ignore the dilation rates while performing our modulated convolution.} $\mathcal{A}^{p}_{d} \in \mathbb{R}^{U \times V}$ at spatial location $p$$=$$(h,w)$ under disparity $d$, the modulated convolution can be formulated as
 \begin{equation}\label{eq:modulateconv}
 y(p,d) = \frac{\sum_{k=1}^{UV} \omega_{k} \cdot \mathcal{A}^{p}_{d}(k) \cdot \Delta m^{p}_{k}}{\sum_{k=1}^{UV} \Delta m^{p}_{k}},
 \end{equation}
 where $y(p)$ denotes the resulting matching cost at spatial location $p$ under disparity $d$, $\omega_{k}$ denotes the weight of our cost constructor at the $k^\textit{th}$ sampling point, and $\Delta m^{p}_{k} \in [0,1]$ is the modulation scalar. Note that, the weights of our cost constructor are shared across different spatial locations and disparity values, while the modulation scalar is spatially varying and only shared among different disparity values.

  With the modulation mechanism, our cost constructor can adjust the contributions of each view at each location to achieve occlusion-aware cost construction. For example, if a scene point is occluded in some views, our OACC will assign small modulation scalars to the occluded pixels to reduce their impact on the matching cost. It is demonstrated in Sec.~\ref{sec:modelanalyses} that the modulation mechanism is crucial for accurate depth estimation.

 \begin{figure}[t]
 \centering
 \includegraphics[width=8.3cm]{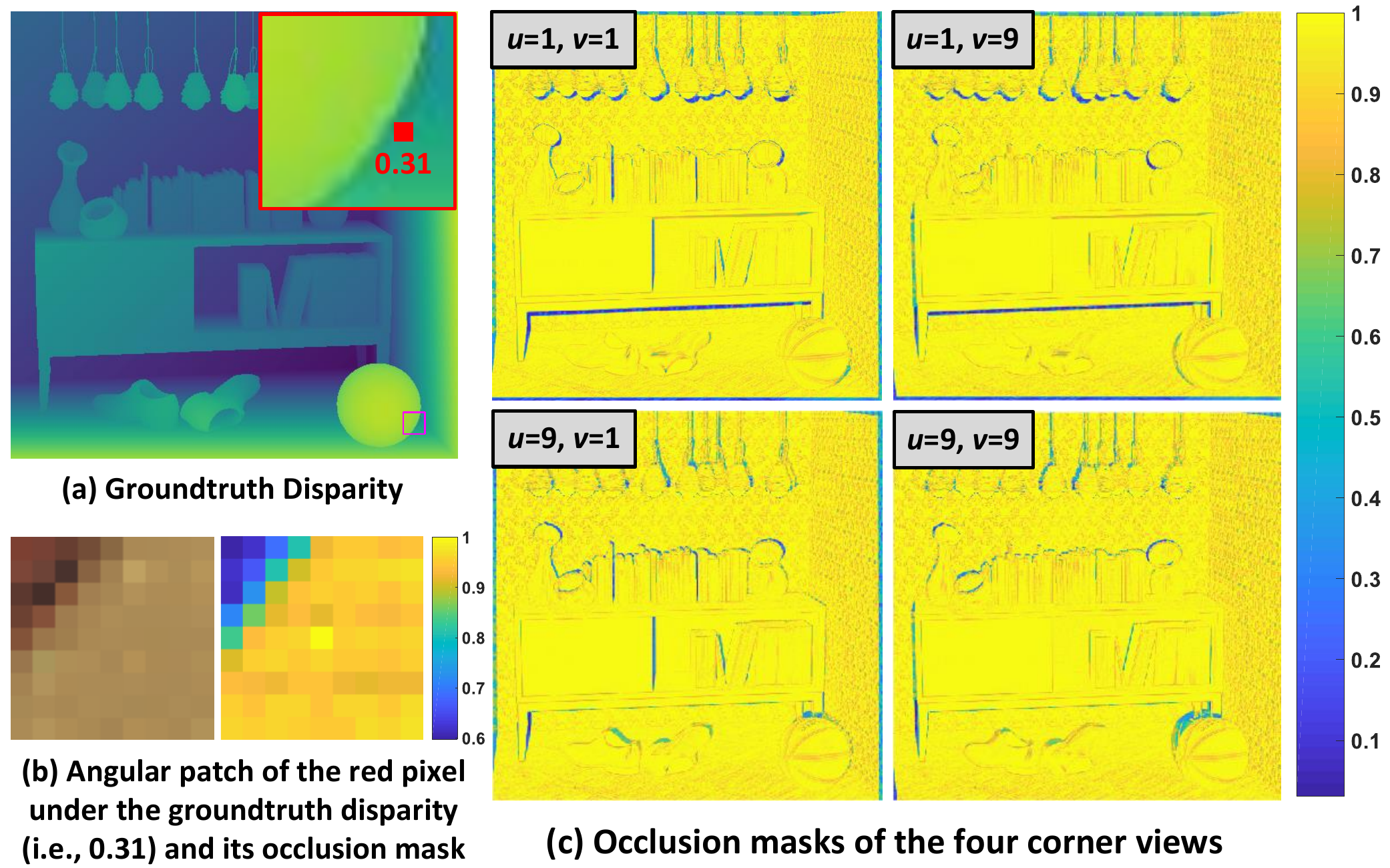}
 \vspace{-0.5cm}
 \caption{Visualization of the generated occlusion masks on scene \textit{sideboard}. Lower values represent heavier occlusions.} \label{fig:OccMask}
 \vspace{-0.3cm}
 \end{figure}

 \subsubsection{Occlusion Mask Generation}\label{sec:OccMaskGeneration}
 To achieve occlusion-aware cost construction, the occlusion mask of each view need to be calculated to generate reasonable modulation scalars in Eq.~\ref{eq:modulateconv}. However, accurate occlusion estimation is a non-trivial task. Inspired by the unsupervised LF depth estimation methods \cite{peng2020zero,jin2021occlusion,zhou2019unsupervised,peng2018unsupervised}, in this paper, we propose a parameter-free approach to deduce occlusion mask of each view. Specifically, for regions with occlusions, a scene point that is available in the center view can be unavailable in the surrounding views, and the occluded pixels in these surrounding views cannot find their corresponding pixels in the center view.  Consequently, the fine-grained occlusion mask can be calculated based on the photometric consistency prior.

 Denote the disparity map of the center view as $\mathcal{D}_{c}$, the surrounding views are firstly warped to the center view, i.e.,
  \begin{equation}
 \mathcal{I}_{k\rightarrow c} = W^{\mathcal{D}_{c}}_{k\rightarrow c}\left(\mathcal{I}_{k}\right), \quad k=1,2,\cdots,UV,
 \end{equation}
 where $W^{\mathcal{D}_{c}}_{k\rightarrow c}$ denotes the warping operation that projects the $k^\textit{th}$ view $\mathcal{I}_{k}$ to the center view $\mathcal{I}_{c}$. Assume that the disparity map $\mathcal{D}_{c}$ is accurate, the projected view $\mathcal{I}_{k\rightarrow c}$ should have identical values to the center view $\mathcal{I}_{c}$ at non-occluded regions. Therefore, we use the absolute residuals between $\mathcal{I}_{k\rightarrow c}$ and $\mathcal{I}_{c}$ to measure the photometric consistency, i.e.,
 \begin{equation}
 \mathcal{I}^{res}_{k\rightarrow c} = \left|\mathcal{I}_{k\rightarrow c} - \mathcal{I}_{c}\right|.
 \end{equation}

 Finally, the occlusion mask of the $k^\textit{th}$ view is obtained by re-mapping $\mathcal{I}^{res}_{k\rightarrow c}$ to $\left[0,1\right]$, i.e.,
 \begin{equation}\label{eq:decay}
 \mathcal{M}_{k} = \left|1 - \mathcal{I}^{res}_{k\rightarrow c}\right|^q,
 \end{equation}
 where $q$ is a scalar that controls the decaying rate. A larger $q$ can enhance the sensitivity to occlusions but degrade the robustness to noise (see Sec~\ref{sec:modelanalyses}). In our implementation, we empirically set $q$$=$$2$ to achieve a good trade-off between occlusion awareness and noise robustness.

 Using the aforementioned approach, the occlusion mask (i.e., the modulation scalars in Eq.~\ref{eq:modulateconv}) of each view can be obtained. As shown in Fig.~\ref{fig:OccMask}, the generated occlusion masks are reasonable and consistent to the real case.

 \begin{figure*}[t]
 \centering
 \includegraphics[width=13.5cm]{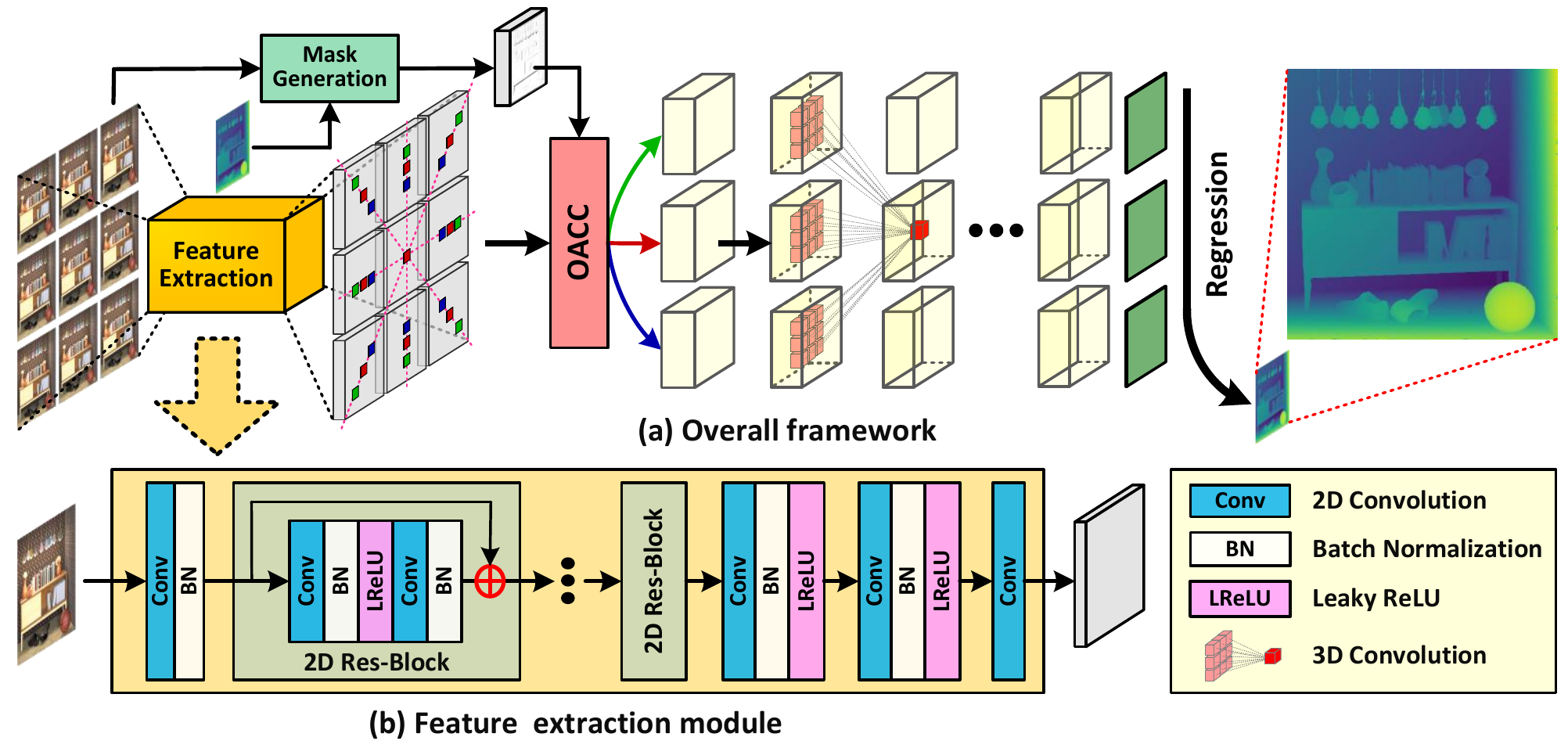}
 \vspace{-0.2cm}
 \caption{An overview of our OACC-Net. Here, a 3$\times$3 LF is used as an example for illustration.} \label{fig:OACC-Net}
 \vspace{-0.1cm}
 \end{figure*}

 \subsection{Network Design} \label{Network Design}
 Based on the proposed OACC, we develop a deep network called OACC-Net for LF depth estimation. As shown in Fig.~\ref{fig:OACC-Net}, our network takes a $U$$\times$$V$ LF as its input and sequentially performs feature extraction, OACC-based cost construction, cost aggregation, and depth regression.

\subsubsection{Feature Extraction}
 As shown in Fig.~\ref{fig:OACC-Net}(b), in our feature extraction module, a 3$\times$3 convolution is first used to extract initial features. Then, eight residual blocks \cite{ResNet} are cascaded for deep feature extraction. These residual blocks are built in a ``Conv-BN-LeakyReLU-Conv-BN'' structure with skip connections for local residual learning. Features generated by the last residual block are further fed to three cascaded 3$\times$3 convolutions to generate features for cost construction. Note that, the weights of all the convolutions in our feature extraction module are shared among different views.

\subsubsection{Cost Construction}\label{sec:costconstruction}
 We use the proposed OACC for matching cost construction. As described in Sec.~\ref{sec:CCviaConv}, features generated by the feature extraction module are organized into SAI arrays to form the input of our OACC. Besides, the occlusion mask of each view is generated for occlusion-aware pixel modulation. However, there is a ``chicken-and-egg'' problem between occlusion mask generation and depth estimation. That is, occlusion mask generation requires a disparity map as its input, but the disparity information is unavailable at this stage.

 Here, we propose an iterative scheme to solve this problem. In the testing phase, we generate an initial occlusion mask by setting all its elements to one (i.e., non-occlusion assumption), and use this initial mask for depth estimation. After obtaining the initial disparity map, we update the occlusion mask and use the updated mask to generate more accurate disparity maps. It is demonstrated in Sec.~\ref{sec:modelanalyses} that the proposed iterative scheme works well and can make occlusion prediction and depth estimation mutually boost. In the training stage, we directly use the groundtruth disparity map to generate occlusion masks to avoid training collapse.

\subsubsection{Cost Aggregation and Regression}
Given the cost volume generated by our OACC, we first use a 1$\times$1 convolution to reduce its channel depth from 512 to 160. Then we cascade eight 3D convolutions with a kernel size of 3$\times$3$\times$3 for cost aggregation. The third to the sixth 3D convolutions are organized into two residual blocks, and channel attention layers are adopted after each residual block to highlight contributive channels. Finally, a 3D tensor $\mathcal{F}_\textit{final}\in \mathbb{R}^{D\times H\times W}$ is generated by the last 3D convolution, and the disparity is regressed according to
\vspace{-0.1cm}
\begin{equation}\label{Eq:regress}
  \hat{\mathcal{D}_{c}} =  \sum^{d_\textit{max}}_{d_{k}=d_\textit{min}} d_{k} \times \textit{Softmax}(\mathcal{F}_\textit{final}),
\end{equation}
\vspace{-0.1cm}
where $\hat{\mathcal{D}_{c}}$ denotes the estimated center-view disparity, $\textit{Softmax}(\cdot)$ denotes the softmax normalization which is performed along the disparity axis of  $\mathcal{F}_\textit{final}$.

\section{Experiments}
In this section, we first introduce the datasets and implementation details, then conduct experiments to investigate our models. Finally, we compare our OACC-Net to several state-of-the-art LF depth estimation methods.

\subsection{Datasets and Implementation Details}
 We used the 4D LF benchmark \cite{HCInew} to validate the effectiveness of our method. All LFs in this benchmark have an angular resolution of 9$\times$9 and a spatial resolution of 512$\times$512. All the 9$\times$9 views are used by our method for depth estimation. We followed \cite{EPINET,LFAttNet,AttMLFNet,FastLFnet} to use 16 scenes in the ``Additional'' category for training, 8 scenes in the ``Stratified'' and ``Training'' categories for validation, and 4 scenes in the ``Test'' category for test. We also used other LF datasets \cite{STFgantry,HCIold,EPFL,INRIA} to test the generalization capability of our method (see Fig.~\ref{fig:Visual-DispReal} and the \href{https://yingqianwang.github.io/files/OACC-Net_supp.pdf}{supplemental material}).

 During the training phase, we randomly cropped SAIs into patches of size 48$\times$48, and converted them into gray-scale images. We performed a large variety of data augmentation, including random flipping and rotation, brightness and contrast adjustment, noise injection, refocusing, and downsampling. Our OACC-Net was trained in a supervised manner with an L1 loss, and was optimized using the Adam method \cite{Adam} with $\beta_1$=0.9, $\beta_2$=0.999. The batch size was set to 16 and the learning rate was set to 1$\times$10$^{-3}$. The training was stopped after 3$\times$10$^{5}$ iterations and takes about 7 days. Our model was implemented in PyTorch and trained on a PC with two Nvidia RTX 2080Ti GPUs.

 We used the mean square error (MSE) and bad pixel ratio (BadPix($\epsilon$)) as quantitative metrics for performance evaluation. BadPix($\epsilon$) measures the percentage of incorrectly estimated pixels whose absolute errors exceeding a predefined threshold (e.g., $\epsilon=$0.07, 0.03, 0.01).

 \subsection{Model Analyses} \label{sec:modelanalyses}
 We first conduct experiment to validate the effectiveness of our pixel modulation mechanism. Then, we test the performance of our method with different number of iterations and decaying rates. Finally, we demonstrate the efficiency of our OACC.

 \begin{figure}
 \centering
 \includegraphics[width=8.3cm]{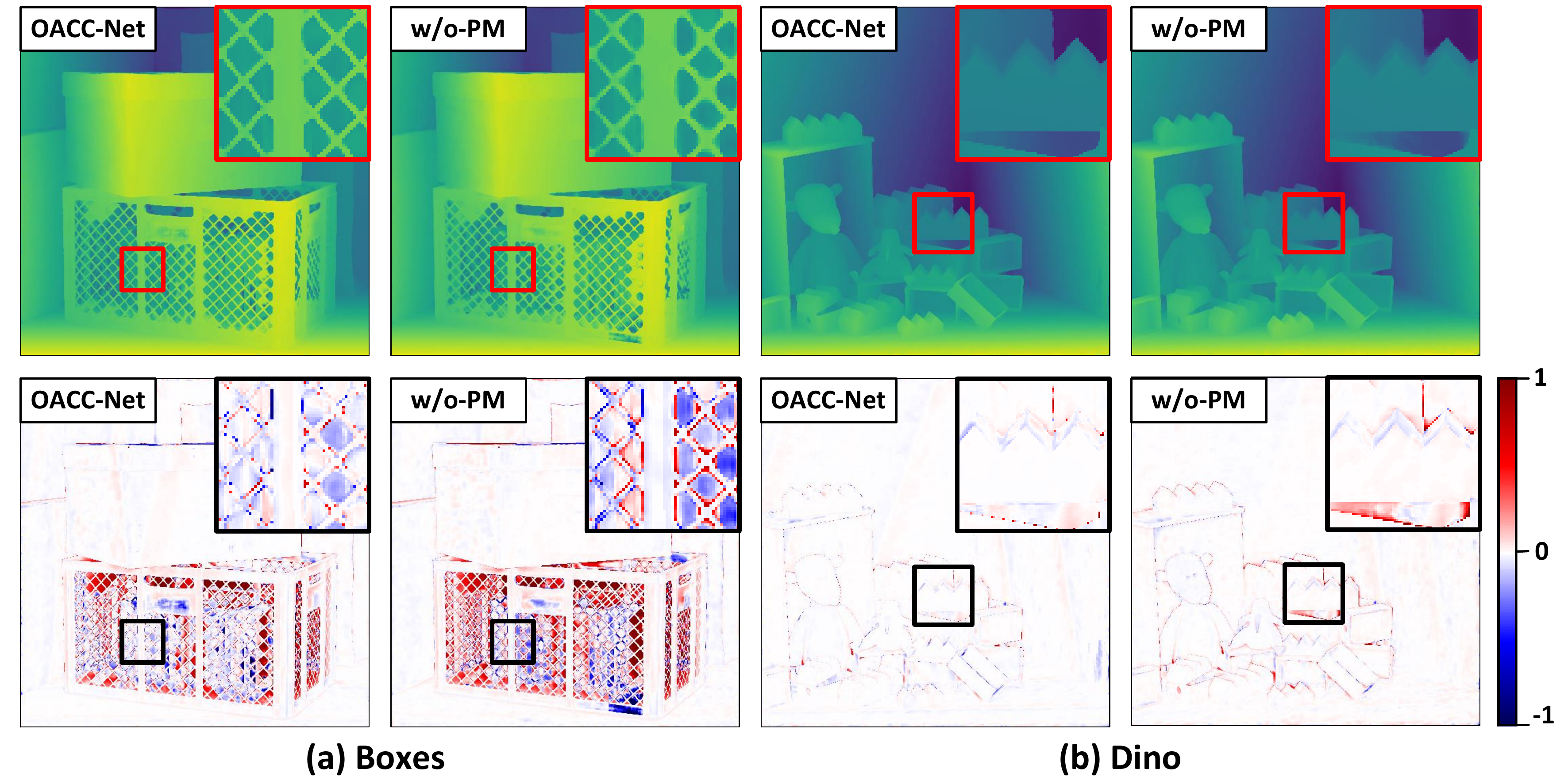}
 \vspace{-0.5cm}
 \caption{Visual comparisons of our method on scenes (a) \textit{boxes} and (b) \textit{dino} with$/$without using pixel modulation mechanism. Top-row figures show the estimated disparity $\hat{\mathcal{D}}$ while the bottom-row figures show the corresponding error maps ($\hat{\mathcal{D}} - \mathcal{D}_\textit{gt}$).} \label{fig:VisualDispMask}
 \vspace{-0.4cm}
 \end{figure}

 \begin{table*}
\caption{BadPix0.07 (BP07) and MSE (multiplied with 100) achieved by different variants of our OACC-Net on the 4D LF benchmark \cite{HCInew}. ``\textit{w/o-PM}'' denotes the model trained without using the pixel modulation mechanism, ``iter\_$k$'' denotes the model performing $k$ iterations in the inference stage, and ``\textit{gt-mask}'' denotes the model using occlusion masks generated by the groundtruth disparities. The main model of our OACC-Net (i.e., \textit{iter\_2}) is highlighted. The best results are in \textcolor{red}{red} and the second best results are in \textcolor{blue}{blue}.}\label{tab:ablation}
\vspace{-0.2cm}
\centering
\scriptsize
\renewcommand\arraystretch{0.95}
\begin{tabular}{|p{1.1cm}<{\centering}|p{0.45cm}<{\centering}p{0.45cm}<{\centering}|p{0.45cm}<{\centering}p{0.45cm}<{\centering}|p{0.45cm}<{\centering}p{0.45cm}<{\centering}|p{0.45cm}<{\centering}p{0.45cm}<{\centering}|p{0.45cm}<{\centering}p{0.45cm}<{\centering}|p{0.45cm}<{\centering}p{0.45cm}<{\centering}|p{0.45cm}<{\centering}p{0.45cm}<{\centering}|p{0.45cm}<{\centering}p{0.45cm}<{\centering}|p{0.45cm}<{\centering}p{0.45cm}<{\centering}|}
\hline
\multirow{2}*{\textit{Models}} & \multicolumn{2}{c|}{\textit{Backgammon}}  &   \multicolumn{2}{c|}{\textit{Dots}} & \multicolumn{2}{c|}{\textit{Pyramids}}  &   \multicolumn{2}{c|}{\textit{Stripes}}  &   \multicolumn{2}{c|}{\textit{Boxes}}  &   \multicolumn{2}{c|}{\textit{Cotton}} & \multicolumn{2}{c|}{\textit{Dino}}  &   \multicolumn{2}{c|}{\textit{Sideboard}} & \multicolumn{2}{c|}{\textbf{Average}}\\
\cline{2-19}
  & \tiny{\textbf{\textit{BP07}}}  & \tiny{\textbf{\textit{MSE}}}
  & \tiny{\textbf{\textit{BP07}}} &  \tiny{\textbf{\textit{MSE}}}
  & \tiny{\textbf{\textit{BP07}}}  & \tiny{\textbf{\textit{MSE}}}
  & \tiny{\textbf{\textit{BP07}}} & \tiny{\textbf{\textit{MSE}}}
  & \tiny{\textbf{\textit{BP07}}}  & \tiny{\textbf{\textit{MSE}}}
  & \tiny{\textbf{\textit{BP07}}} &  \tiny{\textbf{\textit{MSE}}}
  & \tiny{\textbf{\textit{BP07}}}  & \tiny{\textbf{\textit{MSE}}}
  & \tiny{\textbf{\textit{BP07}}} & \tiny{\textbf{\textit{MSE}}}
  & \tiny{\textbf{\textit{BP07}}} & \tiny{\textbf{\textit{MSE}}} \\
\hline
\textit{w/o-PM}
& 7.142  &  5.097 & 2.419  & \textcolor{blue}{1.382}   & 0.269  & 0.008 & 5.909 & 1.084
& 14.17  &  4.023 &  0.550 &  0.174  &  1.649 & 0.133 & 3.846  & 0.674
& 4.494 & 1.572\\

\textit{iter\_1}
& 4.128  &  4.059 & 1.762  & 1.395   & \textcolor{red}{0.156}  & \textcolor{blue}{0.005} & 3.444 & 0.879
& 10.84  &  3.182 &  0.352 &  0.172  &  1.099 & 0.091 & 3.366  & 0.562
& 3.143 & 1.293\\
\rowcolor{shadow}
\textit{iter\_2}
& 3.931  & 3.938  & \textcolor{red}{1.510}  & 1.418  & \textcolor{blue}{0.157} & \textcolor{red}{0.004} & 2.920 & \textcolor{blue}{0.845}
& 10.70  & \textcolor{blue}{2.892}  & 0.312  & 0.162 & \textcolor{blue}{0.967}  & 0.083 & 3.350 & \textcolor{blue}{0.542}
& 2.981 & \textcolor{blue}{1.236} \\
\textit{iter\_3}
& 3.928  & 3.858  & 1.578  & 1.421 & 0.158   & \textcolor{blue}{0.005} & 2.920 & 0.847
& \textcolor{blue}{10.56}  & 2.968  & 0.319  & 0.161 & 0.989   & \textcolor{blue}{0.082} & \textcolor{blue}{3.314} & 0.581
& \textcolor{blue}{2.970} & 1.240 \\
\textit{iter\_4}
& \textcolor{blue}{3.918}  & \textcolor{blue}{3.824}  & 1.572   & 1.424  & 0.158  & \textcolor{blue}{0.005} & \textcolor{blue}{2.909} & 0.846
& 10.60  & 2.966  & \textcolor{blue}{0.310}   &  \textcolor{blue}{0.159} & 0.999  & \textcolor{blue}{0.082} & 3.337 & 0.591
& 2.975 & 1.237 \\
\textit{gt-mask}
& \textcolor{red}{3.910}  & \textcolor{red}{3.593}  & \textcolor{blue}{1.515}   &  \textcolor{red}{1.297} & \textcolor{red}{0.156}  & \textcolor{blue}{0.005} & \textcolor{red}{2.876} & \textcolor{red}{0.840}
& \textcolor{red}{10.01}  & \textcolor{red}{2.278}  & \textcolor{red}{0.300}   &  \textcolor{red}{0.118} & \textcolor{red}{0.909}  & \textcolor{red}{0.072} & \textcolor{red}{3.057} & \textcolor{red}{0.502}
& \textcolor{red}{2.842} & \textcolor{red}{1.088} \\
\hline

\hline
\end{tabular}
\vspace{-0.2cm}
\end{table*}

\textbf{Pixel modulation mechanism.} We replaced the modulated convolution with a vanilla convolution and retrained this variant from scratch. As shown in Table~\ref{tab:ablation}, compared to our OACC-Net (i.e., ``iter\_2''), model ``w/o-PM'' suffers a 1.513 and 0.336 increase in BadPix0.07 and MSE, respectively. That is because, without using the pixel modulation mechanism, pixels from each view and at each spatial location are processed equally thus the occlusion issue cannot be handled. The qualitative results in Fig.~\ref{fig:VisualDispMask} also demonstrate that the disparities predicted by our OACC-Net are more accurate at regions with heavy occlusions.

\textbf{Number of iterations.} We compare the performance of our method with different number of iterations (see Sec \ref{sec:costconstruction}). As shown in Table~\ref{tab:ablation}, directly using an all-one tensor as an occlusion mask (i.e., ``iter\_1'') achieves 3.143 and 1.293 in terms of average BadPix0.07 and average MSE, respectively. It demonstrates the importance of occlusion mask generation in our method. When using the initial estimated disparity to generate occlusion masks, the average BadPix0.07 and MSE values are reduced to 2.981 and 1.236, respectively, which demonstrates the effectiveness of our iterative scheme. Note that, performing more iterations (i.e., ``iter\_3'' and ``iter\_4'') cannot introduce significant improvements. Therefore, we set the iteration number to 2 in our OACC-Net for a good trade-off between accuracy and efficiency. Moreover,  we explore the upper bound of our method by using the occlusion mask generated by groundtruth disparity. As shown in Table~\ref{tab:ablation}, by using ``groundtruth'' masks, occlusions can be well located and the accuracy is improved. It demonstrates that accurate occlusion masks are important for LF depth estimation.

\begin{figure}
 \centering
 \includegraphics[width=8.3cm]{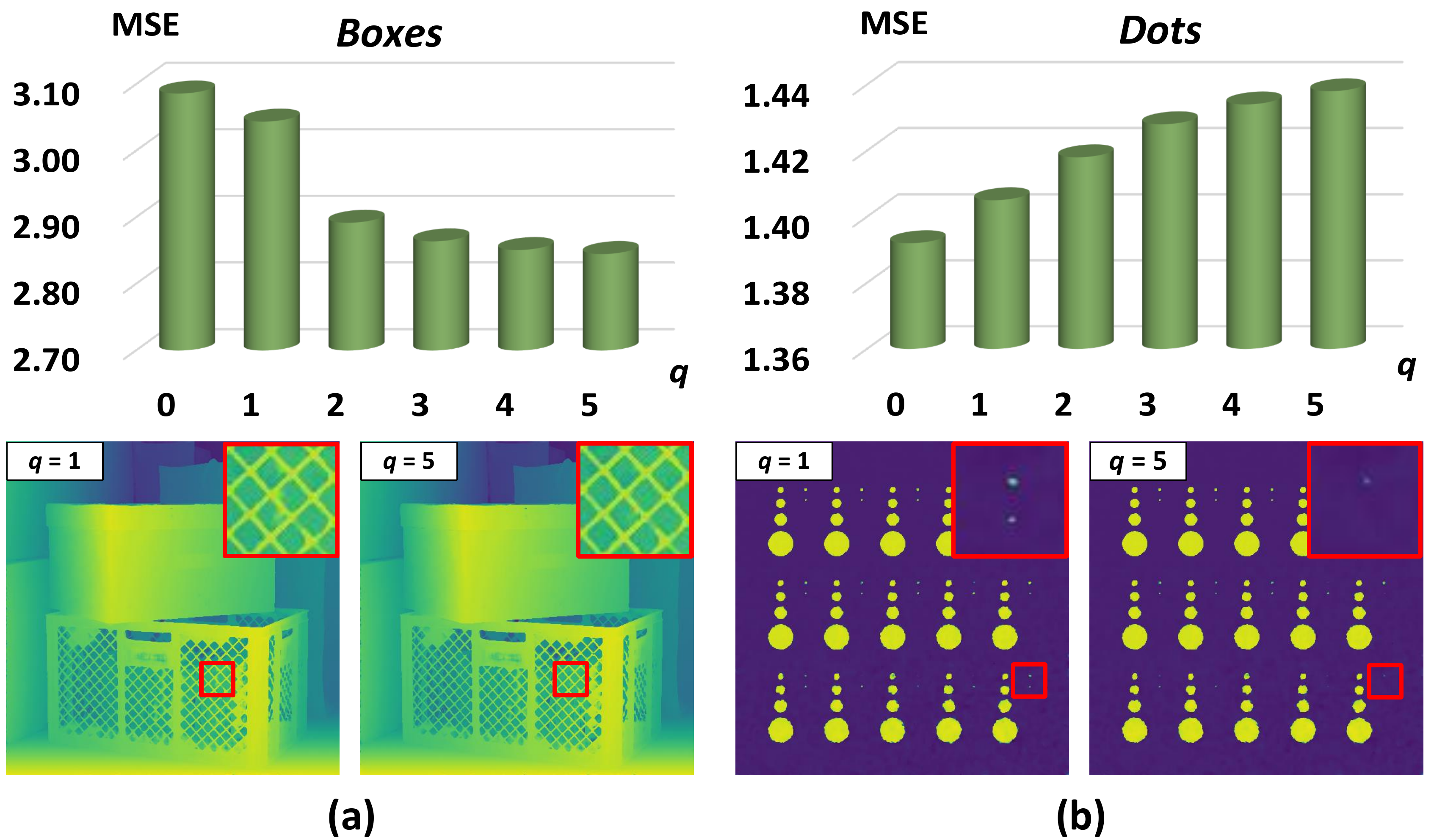}
 \vspace{-0.6cm}
 \caption{Results achieved by our method on scenes (a) \textit{boxes} and (b) \textit{dots} with different decaying rates $q$. Top figures show the MSE \textit{w.r.t.} different decaying rates, and the bottom figures show the disparity maps with $q$$=$$1$ and $q$$=$$5$.} \label{fig:MSE-q}
 \vspace{-0.4cm}
 \end{figure}

\textbf{Effect of decaying rate.} We compare our method with different decaying rates in Eq.~\ref{eq:decay}. As shown in Fig.~\ref{fig:MSE-q}, larger decaying rates can make our method perform better on scenes with heavy occlusions (see Fig.~\ref{fig:MSE-q}(a)) but reduce the robustness to large noise (see Fig.~\ref{fig:MSE-q}(b)). That is because, under large noise, the photometric consistency can be broken and some pixels can be mis-classified as occluded pixels with large decaying rates. Consequently, we set $q$$=$$2$ in our method for a good trade-off between occlusion-awareness and noise-robustness.

\begin{table}
\caption{Model size and running time of our OACC-Net at different stages. Here, a 9$\times$9 LF with a spatial resolution of 512$\times$512 is used as input. The proposed OACC achieves a very fast inference speed and significantly accelerates our OACC-Net.}\label{tab:runningtime}
\vspace{-0.2cm}
\scriptsize
\renewcommand\arraystretch{1.0}
\centering
\begin{tabular}{|c|c|c|}
\hline
 \multirow{2}*{Stages} & \multirow{2}*{\textit{Shift-and-Concat}} & \multirow{2}*{\textit{OACC-Net} (ours)} \\
 &&\\
 \hline
  feature extraction  & 0.04 M $/$ 0.004 s & 0.04 M $/$ 0.004 s \\
  \textbf{cost construction} &  ~~~ 0 M $/$ \textbf{2.741} s  &  0.04 M $/$ \textbf{0.004} s \\
  aggregation \& regression & 4.93 M $/$ 0.003 s & 4.93 M $/$ 0.003 s \\
  mask generation   & - & ~~~ 0 M $/$ 0.015 s \\
  cost construction &  -  &  shared $/$ 0.004 s \\
  aggregation \& regression & - & shared $/$ 0.004 s \\
  \hline
  Total     & 4.97 M $/$ 2.748 s & 5.01 M $/$ 0.034 s \\
\hline
\end{tabular}
\vspace{-0.2cm}
\end{table}

\textbf{Efficiency.}  We investigate the efficiency of our method by listing the model size and running time of our OACC-Net at each stage. Here, we introduce a variant by using the \textit{shift-and-concat} approach for cost construction, where 80 views are shifted by 8 disparity levels. As shown in Table~\ref{tab:runningtime}, the \textit{shift-and-concat} approach spends 2.741 seconds on cost construction while our OACC spends only 4 milliseconds. Since our OACC can directly convolve pixels under specific disparities and avoids the repetitive shifting operation, the inference speed of our OACC-Net is significantly accelerated at the cost of a 0.04 M increase in model size.

\begin{table*}
\caption{
Mean square error (multiplied with 100) and average running time achieved by different methods on the 4D LF benchmark \cite{HCInew}. The best results are in \textcolor{red}{red} and the second best results are in \textcolor{blue}{blue}.}\label{tab:QuantitativeDisp}
\vspace{-0.2cm}
\centering
\scriptsize
\renewcommand\arraystretch{0.95}
\begin{tabular}{|l|
p{0.62cm}<{\centering}p{0.62cm}<{\centering}p{0.62cm}<{\centering}p{0.62cm}<{\centering} p{0.62cm}<{\centering}p{0.62cm}<{\centering}p{0.62cm}<{\centering}p{0.62cm}<{\centering}  p{0.62cm}<{\centering}p{0.62cm}<{\centering}p{0.62cm}<{\centering}p{0.62cm}<{\centering}|p{0.62cm}<{\centering}|p{0.62cm}<{\centering}|}
\hline
\hline

 \tiny{\textbf{\textit{Method}}} & \tiny{\textbf{\textit{Backgm}}} &  \tiny{\textbf{\textit{Dots}}} & \tiny{\textbf{\textit{Pyramids}}} & \tiny{\textbf{\textit{Strips}}} &
 \tiny{\textbf{\textit{Boxes}}} &  \tiny{\textbf{\textit{Cotton}}} &  \tiny{\textbf{\textit{Dino}}} & \tiny{\textbf{\textit{Sideboard}}} &
 \tiny{\textbf{\textit{Bedroom}}} &  \tiny{\textbf{\textit{Bicycle}}} &  \tiny{\textbf{\textit{Herbs}}} & \tiny{\textbf{\textit{Origami}}} & \tiny{\textbf{\textit{Average}}} & \tiny{\textbf{\textit{Time}} (s)}\\
\hline
\textit{LF\_OCC} \cite{LF-OCC} &
22.78 & 3.185 & 0.077 & 7.942 &
9.593 & 1.074 & 0.944 & 2.073 &
0.530 & 7.673 & 22.96 & 2.223 & 6.755 & 519.9 \\

\textit{CAE} \cite{CAE} &
6.074 & 5.082 & 0.048 & 3.556 &
8.424 & 1.506 & 0.382 & 0.876 &
0.234 & 5.135 & 11.67 & 1.778 & 3.730 & 832.1 \\

\textit{PS-RF} \cite{PS-RF} &
6.892 & 8.338 & 0.043 & 1.382 &
9.043 & 1.161 & 0.751 & 1.945 &
0.288 & 7.926 & 15.25 & 2.393 & 4.617 & 1413 \\

\textit{SPO} \cite{SPO} &
4.587 & 5.238 & 0.043 & 6.955 &
9.107 & 1.313 & 0.310 & 1.024 &
0.209 & 5.570 & 11.23 & 2.032 & 3.968 & 2115 \\

\textit{SPO-MO} \cite{SPO-MO} &
4.133 & 3.763 & 0.009 & 1.934 &
10.37 & 1.329 & 0.254 & 0.932 &
\textcolor{blue}{0.152} & 5.617 & 12.05 & 1.667 & 3.518 & 4304 \\

\textit{OBER-cross-ANP} \cite{OBER} &
4.700 & 1.757 & 0.008 & 1.435 &
4.750 & 0.555 & 0.336 & 0.941 &
0.185 & 4.314 & 10.44 & 1.493 & 2.584 & 183.0 \\

\textit{OAVC} \cite{OAVC} &
3.835 & 16.58 & 0.040 & 1.316 &
6.988 & 0.598 & 0.267 & 1.047 &
0.212 & 4.886 & 10.36 & 1.478 & 3.968 & 19.41 \\

\textit{EPN+OS+GC} \cite{EPN} &
3.699 & 22.37 & 0.018 & 8.731 &
9.314 & 1.406 & 0.565 & 1.744 &
1.188 & 6.411 & 11.58 & 10.09 & 6.426 & 274.7 \\

\textit{Epinet-fcn} \cite{EPINET} &
\textcolor{blue}{3.629} & 1.635 & 0.008 & 0.950 &
6.240 & \textcolor{blue}{0.191} & 0.167 & 0.827 &
0.213 & 4.682 & 9.700 & \textcolor{blue}{1.466} & 2.476 & 1.976 \\

\textit{Epinet-fcn-m} \cite{EPINET} &
3.705 & 1.475 & \textcolor{blue}{0.007} & 0.932 &
5.968 & 0.197 & 0.157 & 0.798 &
0.204 & 4.603 & 9.491 & 1.478 & 2.418 & 10.66 \\

\textit{Epinet-fcn-9x9} \cite{EPINET} &
3.909 & 1.980 & \textcolor{blue}{0.007} & 0.915 &
6.036 & 0.223 & 0.151 & 0.806 &
0.231 & 4.929 & 9.423 & 1.646 & 2.521 & 2.041 \\

\textit{EPI-Shift} \cite{EPI-Shift} &
12.79 & 13.15 & 0.037 & 1.686 &
9.790 & 0.475 & 0.392 & 1.261 &
0.284 & 6.920 & 17.01 & 1.690 & 5.458 & 22.57 \\

\textit{EPI\_ORM} \cite{ORM} &
\textcolor{red}{3.411} & 14.48 & 0.016 & 1.744 &
4.189 & 0.287 & 0.336 & 0.778 &
0.298 & 3.489 & \textcolor{red}{4.468} & 1.826 & 2.944 & 76.61 \\

\textit{LFAttNet} \cite{LFAttNet} &
3.648 & \textcolor{blue}{1.425} & \textcolor{red}{0.004} & \textcolor{blue}{0.892} &
\textcolor{blue}{3.996} & 0.209 & \textcolor{blue}{0.093} & \textcolor{red}{0.530} &
0.366 & \textcolor{blue}{3.350} & 6.605 & 1.733 & 1.904 & 5.862 \\

\textit{FastLFnet} \cite{FastLFnet} &
3.986 & 3.407 & 0.018 & 0.984 &
4.395 & 0.322 & 0.189 & 0.747 &
0.202 & 4.715 & 8.285 & 2.228  & 2.456 & \textcolor{blue}{0.624} \\

\textit{OACC-Net} (ours) &
3.938 & \textcolor{red}{1.418} & \textcolor{red}{0.004} & \textcolor{red}{0.845} &
\textcolor{red}{2.892} & \textcolor{red}{0.162} & \textcolor{red}{0.083} & \textcolor{blue}{0.542} &
\textcolor{red}{0.148} & \textcolor{red}{2.907} & \textcolor{blue}{6.561} & \textcolor{red}{0.878} & \textcolor{red}{1.698} & \textcolor{red}{0.034} \\
\hline

\end{tabular}
\vspace{-0.1cm}
\end{table*}

 \begin{figure*}[t]
 \centering
 \includegraphics[width=17.3cm]{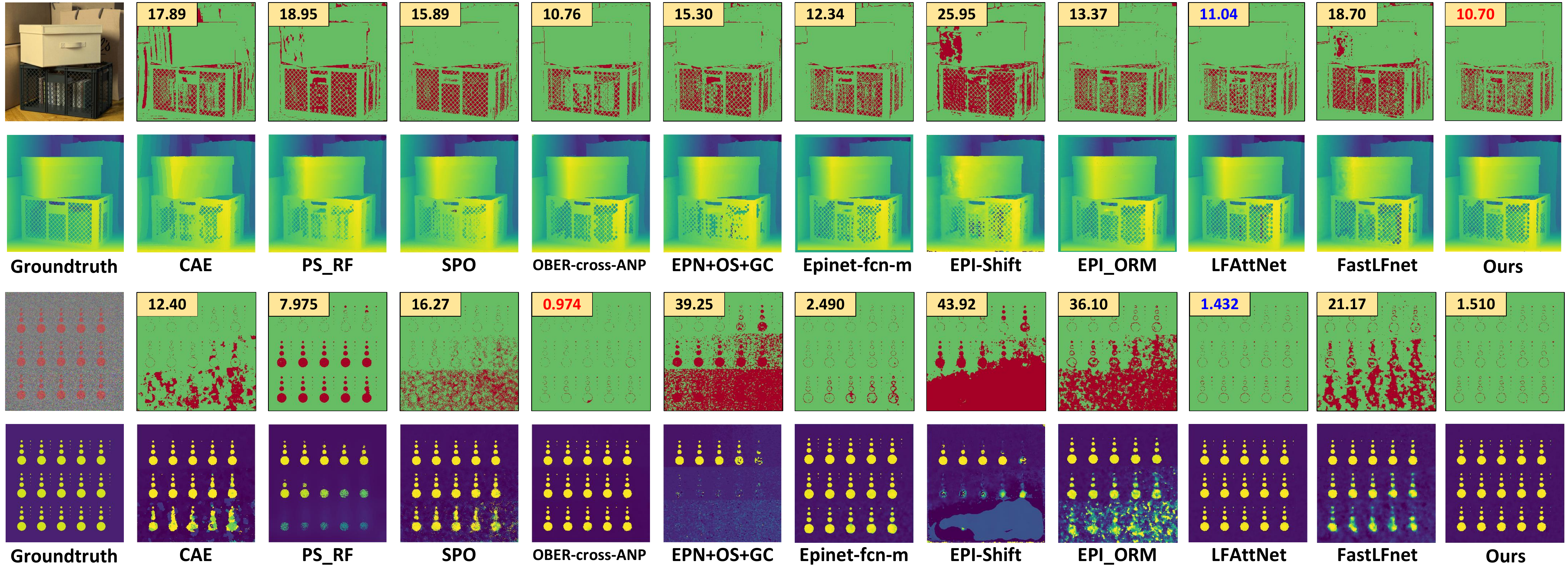}
 \vspace{-0.2cm}
 \caption{Visual comparisons among different LF depth estimation methods on the 4D LF benchmark \cite{HCInew}. For each scene, the bottom row shows the estimated disparity maps and the top row shows the corresponding BadPix0.07 maps (pixels with absolute error larger than 0.07 are marked in red).} \label{fig:Visual-Disp}
 \vspace{-0.3cm}
 \end{figure*}

\subsection{Comparison to the State-of-the-art Methods}
We compare our method to 13 state-of-the-art methods, including 7 traditional methods \cite{LF-OCC,CAE,PS-RF,SPO,SPO-MO,OBER,OAVC} and 6 deep learning-based methods \cite{EPN,EPINET,EPI-Shift,ORM,LFAttNet,FastLFnet}.

\textbf{1) Quantitative Results:} Table \ref{tab:QuantitativeDisp} shows the MSE and average running time achieved by different methods. It can be observed that our method achieves the lowest MSE (i.e., highest accuracy) on 9 scenes and the second lowest MSE on 2 scenes. We submitted our results to the 4D LF benchmark \cite{HCInew} for a comprehensive evaluation. Among all the 91 submissions, our method achieves the first and the second place in terms of average MSE and average BadPix0.07, respectively. Readers can refer to our \href{https://yingqianwang.github.io/files/OACC-Net_supp.pdf}{supplemental material} for additional results.
Note that, our method only spends 0.034 seconds on each scene, which is faster than FastLFnet \cite{FastLFnet} by an order of magnitude. The high accuracy and efficiency demonstrate the superiority of our OACC.

\begin{figure}
 \centering
 \includegraphics[width=8cm]{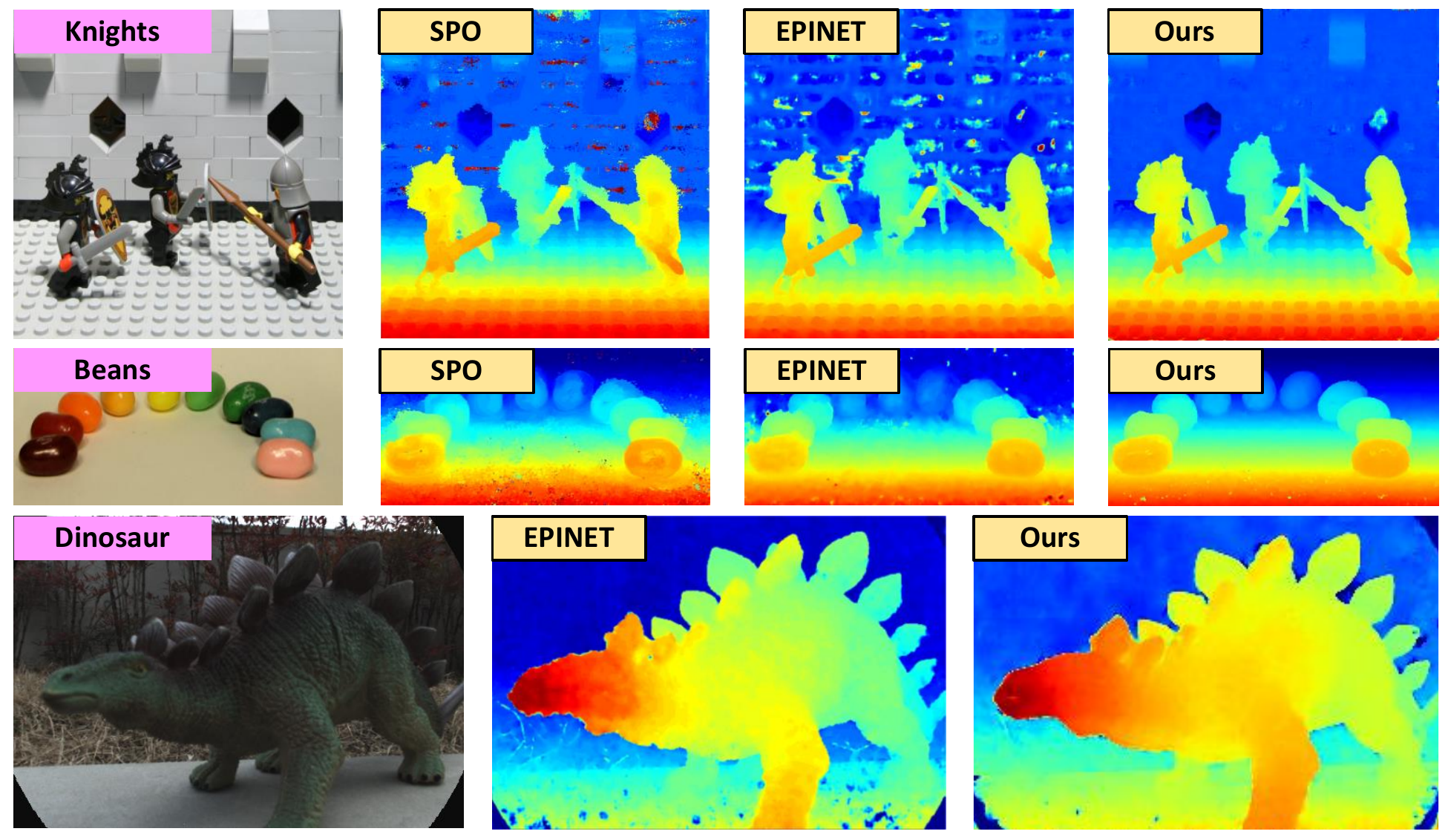}
 \vspace{-0.3cm}
 \caption{Visual results achieved by SPO \cite{SPO}, EPINET \cite{EPINET}, and our method on real LFs.} \label{fig:Visual-DispReal}
 \vspace{-0.5cm}
 \end{figure}

\textbf{2) Visual Comparison:} Figure~\ref{fig:Visual-Disp} shows the estimated disparities and corresponding BadPix0.07 maps. Since the proposed OACC can handle occlusions in a fine-grained manner, our OACC-Net performs well on scenes with heavy and complex occlusions (e.g., the nested structures in scene \textit{boxes}). Besides, our method is also robust to noise and outperforms many state-of-the-art methods \cite{FastLFnet,EPINET,ORM,EPI-Shift} on scenes with large noise (e.g., the bottom dots in scene \textit{dots}).

\textbf{3) Performance on real LFs.}
We test the performance of our OACC-Net on real LFs captured by a moving camera \cite{STFgantry} and a Lytro camera \cite{bok2016geometric}. Since groundtruth depths are unavailable, we used the model trained on the synthetic LFs \cite{HCInew} for inference and compare the visual performance of our method to SPO \cite{SPO} and EPINET \cite{EPINET}. As shown in Fig.~\ref{fig:Visual-DispReal}, the depth maps produced by our method are more reasonable with fewer artifacts. It demonstrates that our OACC-Net can well generalize to real LFs. Please refer to our \href{https://yingqianwang.github.io/files/OACC-Net_supp.pdf}{supplemental material} for additional comparisons.

 \section{Conclusion}
 In this paper, we proposed an occlusion-aware cost constructor for LF depth estimation. Our OACC can efficiently construct matching cost and handle occlusions by modulating input pixels. Based on OACC, we developed a deep network called OACC-Net for depth estimation. Our method is highly efficient and achieves better performance than many state-of-the-art methods on different scenarios.

{\small
\bibliographystyle{ieee_fullname}
\bibliography{ref}
}

\end{document}